\pdfoutput=1

\documentclass[11pt]{article}

\usepackage[]{acl}

\usepackage{times}
\usepackage{latexsym}

\usepackage[T1]{fontenc}

\usepackage[utf8]{inputenc}

\usepackage{microtype}

\usepackage{inconsolata}
\usepackage{graphicx}
\usepackage{placeins}
\usepackage{float}

\usepackage{multirow}

\newcommand{\cfiltlogo}{\raisebox{3.4pt}{\includegraphics[scale=0.026]{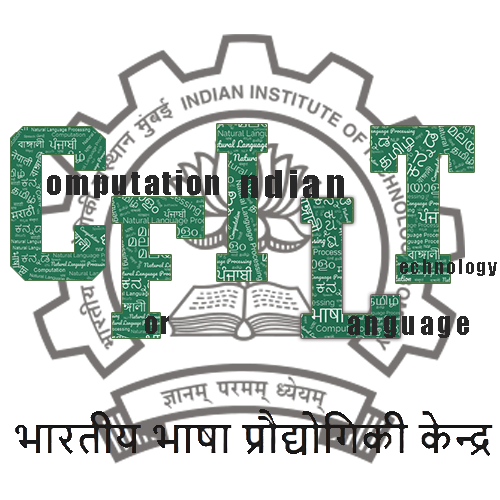}}}
\newcommand{\PAIlogo}{\raisebox{3.4pt}{\includegraphics[scale=0.09]{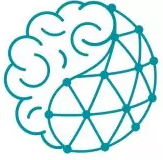}}}

\title{Together We Can: Multilingual Automatic Post-Editing\\ for Low-Resource Languages}

\newcommand*{\affaddr}[1]{#1} 

\newcommand*{\email}[1]{\texttt{#1}}

\author{%
Sourabh Deoghare\cfiltlogo, Diptesh Kanojia\PAIlogo and Pushpak Bhattacharyya\cfiltlogo\\
\affaddr{\cfiltlogo CFILT, Indian Institute of Technology Bombay, Mumbai, India}\\
\affaddr{\PAIlogo Institute for People-Centred AI, University of Surrey, United Kingdom}\\[.2em]
\email{\{sourabhdeoghare, pb\}@cse.iitb.ac.in}, \email{d.kanojia@surrey.ac.uk} \\
}

\begin{document}
\maketitle

\begin{abstract}
This exploratory study investigates the potential of multilingual Automatic Post-Editing (APE) systems to enhance the quality of machine translations for low-resource Indo-Aryan languages. Focusing on two closely related language pairs, English-Marathi and English-Hindi, we exploit the linguistic similarities to develop a robust multilingual APE model.
To facilitate cross-linguistic transfer, we generate synthetic Hindi-Marathi and Marathi-Hindi APE triplets.
Additionally, we incorporate a Quality Estimation (QE)-APE multi-task learning framework. While the experimental results underline the complementary nature of APE and QE, we also observe that QE-APE multitask learning facilitates effective domain adaptation. Our experiments demonstrate that the multilingual APE models outperform their corresponding English-Hindi and English-Marathi single-pair models by $2.5$ and $2.39$ TER points, respectively, with further notable improvements over the multilingual APE model observed through multi-task learning ($+1.29$ and $+1.44$ TER points), data augmentation ($+0.53$ and $+0.45$ TER points) and domain adaptation ($+0.35$ and $+0.45$ TER points). We release the synthetic data, code, and models accrued during this study publicly\footnote{\href{https://github.com/cfiltnlp/Multilingual-APE}{Github: Multilingual-APE}}.

\end{abstract}

\section{Introduction}
Automatic Post-Editing (APE) aims to address the limitations of an unknown machine translation (MT) system by automatically correcting recurrent translation errors, thereby enhancing the overall translation output~\cite{bojar-etal-2015-findings}. WMT has been driving the research in this field through APE-shared tasks over the years \cite{bojar-etal-2016-findings, chatterjee-etal-2018-findings, chatterjee-etal-2020-findings, bhattacharyya-etal-2023-findings}. However, as the generation of authentic (human-annotated) APE data is difficult, publicly available APE data is limited in terms of language pairs and sizes. The unavailability of an adequate amount of APE data for closely related languages has resulted in research focusing on different APE models for different language pairs.

Therefore, while we see considerable research devoted to advancing Neural Machine Translation (NMT) systems through multilingual MT \cite{johnson2017google, aharoni-etal-2019-massively, dabre2020survey, costa2022no, gala2023indictrans}, to the best of our knowledge, multilingual APE (MAPE) has not studied yet. Akin to how using multiple languages together helps improve translation capabilities, extending this approach to the APE task should enhance the translation correction capabilities for closely related languages despite limited resources.

This study focuses on English-Hindi (\textbf{En-Hi}) and English-Marathi (\textbf{En-Mr}) APE. Hindi and Marathi are two prominent Indo-Aryan languages with notable linguistic similarities because of their common linguistic ancestry \cite{chatterji1969indo, masica1993indo}. For example, both languages utilize the Devanagari script for writing and exhibit substantial vocabulary overlap~\cite{kanojia-etal-2020-challenge,kanojia-etal-2020-harnessing}, largely due to their shared roots in Sanskrit. Grammatically, they both follow the Subject-Object-Verb (SOV) sentence structure, providing a familiar syntactic framework for speakers of either language~\cite{subbarao2012south}.

Such linguistic similarities enable cross-linguistic transfer and suggest that advancements in Multilingual APE (\textbf{MAPE}) could be particularly beneficial for improving translation quality across both these language pairs. Our contributions are:
\begin{enumerate}
    \item \textit{A multilingual APE framework} with data augmentation technique for low-resource language pairs with closely related target languages. The 
    MAPE model trained on English-Hindi and English-Marathi pairs outperforms their corresponding single-pair APE models by $3.03$ and $2.84$ TER points, respectively  (Refer Table \ref{tab:res1}, Table \ref{tab:res2}).
    \item An extension of the \textit{multitask-learning-based APE-QE framework} to MAPE. A multilingual model trained on En-Hi and En-Mr APE and QE tasks surpasses the performance of the single-task-based MAPE model by $1.29$ and $1.44$ TER points for En-Hi and En-Mr pairs, respectively (Refer Table \ref{tab:res3}).
    \item A multitask learning-based \textit{domain-adaptation technique} to cater to resource-constrained settings. Fine-tuning the MAPE model in the multitask-learning fashion only for the domain of interest improves overall performance over the non-adapted MAPE model by $0.35$ and $0.45$ TER points (Refer Table \ref{tab:res5}).
\end{enumerate}


\section{Related Work}
WMT has been driving the APE research over the years through the APE shared tasks \cite{akhbardeh-etal-2021-findings, bhattacharyya-etal-2022-findings, bhattacharyya-etal-2023-findings}. We see a paradigm shift from the WMT18 APE shared task onwards, where supervised transformed-based APE approaches became prominent for the task of improving high-quality translations obtained from neural MT systems \cite{chatterjee-etal-2018-findings}.

Most neural APE approaches leverage transfer learning through the use of pre-trained language or translation models in APE model training \cite{lopes-etal-2019-unbabels, wei-etal-2020-hw, sharma-etal-2021-adapting, deoghare-bhattacharyya-2022-iit}. To obtain latent representations of source and target sentences, \citet{lee-etal-2020-postech} employ multilingual or cross-lingual models. While these approaches followed a two-step training approach where the models are initially trained on synthetic APE data and then on authentic APE data, \citet{oh-etal-2021-netmarble} demonstrated an increasingly complex multi-step training approach, referred to as the Curriculum Training Strategy (CTS), under which a model is gradually trained on more and more advanced tasks enhances APE performance. \citet{yang-etal-2020-hw, deoghare-bhattacharyya-2022-iit} demonstrated that increasing APE data with external MT candidates or through phrase-level APE triplets boosts feature diversity. Additionally, \citet{yang-etal-2020-hw, huang-etal-2022-luls} showed that incorporating domain information into the decoder adapter layers aids in domain adaptation of APE outputs. While we see the use of multilingual pre-trained models, the aim has always been to develop a single-pair APE model.

We also see some work in the literature on the joint exploration of APE and QE. ~\citet{martins-etal-2017-pushing} used APE for the QE improvement.~\citet{hokamp-2017-ensembling} used an ensemble of factored NMT models for word-level QE and APE tasks. ~\citet{chatterjee-etal-2018-combining, deoghare-etal-2023-quality} compared multiple QE-APE coupling techniques and explored the extent of the complimentary nature of these two tasks. \citet{deoghare-etal-2023-quality} proposed a QE-APE multitask learning framework to train a single model on QE and APE tasks jointly through a sophisticated multitask learning approach, Nash-MTL \cite{Navon2022MultiTaskLA}.


\begin{figure}[h]
\centering
\includegraphics[width=\linewidth]{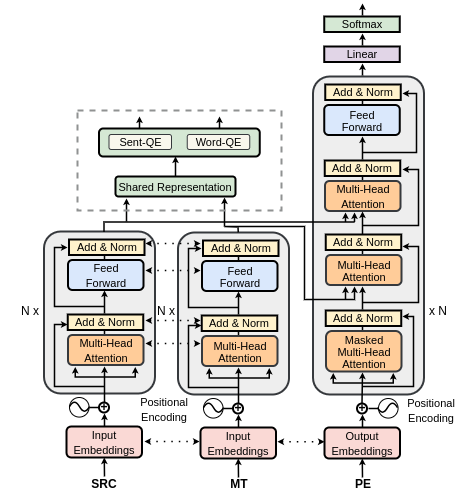}
\caption{APE model architecture \cite{deoghare-etal-2023-quality}}
\label{fig:ape_arch}
\end{figure}

\section{Methodology}

Neural APE approaches model the APE problem as a multi-source translation task. For each pair of the input source sentence and its machine-generated translation as $x=\left\{ x_i \right\}_{i=1}^{T_x}$, $z=\left\{ z_i \right\}_{i=1}^{T_z}$, with lengths $T_x$, and $T_z$, respectively, the APE is trained in a supervised fashion to generate the post-edited translation $\hat{y}=\left\{\hat{y}_i \right\}_{i=1}^{T_{\hat{y}}}$ based on the reference post-edit $y=\left\{ y_i \right\}_{i=1}^{T_y}$ of length $T_y$. For training the model, the cross-entropy loss is used as shown in Equation~\ref{eq:ape_loss}. 

\begin{equation}
\label{eq:ape_loss}
L_{APE} = -\sum_{w=1}^{|S|}\sum_{e=1}^{|V|}y_{w,e}\log\left( \hat{y}_{w,e} \right)
\end{equation}

Where $|S|$ and $|V|$ denote the length of the sequence and the vocabulary size, respectively. $\hat{y}_{w,e}$, and $y_{w,e}$ represent the APE output and its reference, respectively.


While APE aims to generate corrected translations, QE systems are focused on assessing the \textit{extent of correction} required (Sentence-level QE) and precise identification of \textit{locations where those corrections are required} (Word-level QE). The task-specific head is added to the representations obtained from the final encoder layer to perform these tasks.

The Sentence-level QE task is modeled as a regression task with the goal of predicting the Direct Assessment (DA) score for given source sentences and their translation. Equation~\ref{eq:sent_loss} shows the Mean Squared Error (MSE) loss used for this task.

\begin{equation}
\label{eq:sent_loss}
    \mathcal{L}_{sent} = MSE\Big(y_{da}, \hat{y}_{da} \Big)
 \end{equation}

 Where $\hat{y}_{da}$ and $y_{da}$ are the predicted DA score and the ground truth, respectively.

 The Word-level Quality Estimation task is treated as a token-level classification task in which each token is classified into OK/BAD tags to denote its correctness. The cross-entropy loss utilized for this task is shown in Equation \ref{eq:word_loss}.

\begin{equation}
\label{eq:word_loss}
     \mathcal{L}_{word} = -\sum^2_{i=1} \Big(y_{word} \odot \log( \hat{y}_{word} ) \Big)[i]
\end{equation}

Where $y_{word}$ and $\hat{y}_{word}$ denote ground truths and predicted tags.

This complementary nature of APE and QE tasks has been exploited in the QE-APE multitask learning-based framework proposed by \citet{deoghare-etal-2023-quality} in which a two-encoder single-decoder model is jointly trained on APE, sentence-level QE, and word-level QE tasks. They observed higher APE performance when a sophisticated multitask learning algorithm, referred to as Nash-MTL \cite{Navon2022MultiTaskLA}, is used instead of linear scalarization where the task-specific losses are simply added together as shown in Equation \ref{eq:mtl_ls_loss}. We use the Nash-MTL approach as described in \citet{deoghare-etal-2023-quality}.

\begin{equation}
\label{eq:mtl_ls_loss}
L_{LS-MTL} = L_{sent} + L_{word} + L_{APE}
\end{equation}

\paragraph{Model Architecture} In our work, we follow the approach proposed by \citet{deoghare-etal-2023-quality} for training the APE models and also use the same model architecture as shown in Figure \ref{fig:ape_arch}. The architecture without the components shown in the dashed rectangle illustrates the architecture of the non-multitask learning-based APE model before the fine-tuning stage. When fine-tuning, no changes are made to the architecture in the case of single-task experiments. However, the QE task-specific heads are added to the shared representation layer that receives inputs from the final encoder layers for the multitask learning-based experiments.

Except for the transfer learning-based experiments, the encoder weights are initialized using IndicBERT~\cite{kakwani-etal-2020-indicnlpsuite}, and the decoder weights are initialized randomly. We train all models using the Curriculum Training Strategy (CTS), same as \citet{deoghare-etal-2023-quality}.

We consider the following three \textbf{baselines} in our work.

\paragraph{Do Nothing} This baseline treats the original machine translations passed to the model as APE outputs.

\paragraph{Baseline APE} In this experiment, we train separate APE models for English-Hindi and English-Marathi pairs. We consider these APE systems as \textbf{primary baselines}.

\paragraph{Transfer Learning} To train an APE model for one language pair, we use the trained APE checkpoint of the other language pair. For example, to train an En-Mr APE model, we use the \textit{En-Hi Baseline APE} and vice versa.

\section{Multilingual Automatic Post-Editing}

In the hope of getting help from the shared linguistic features of Hindi and Marathi, we perform various MAPE experiments hierarchically.

We leverage synthetic and authentic APE triplets of both En-Hi and En-Mr language pairs for all MAPE experiments. We combine and shuffle the synthetic APE datasets of both language pairs to create \textit{\textbf{multilingual synthetic APE dataset}}. Similarly, we combine and shuffle the authentic APE datasets of both pairs to create \textit{\textbf{multilingual authentic APE dataset}}. These combined sets of multilingual synthetic and authentic APE datasets are then used to train a MAPE model.

As an \textbf{initial approach}, we do not provide the model any indication about the language ID on the target side and represent it as \textbf{w/o-LangID} in further discussion.

\paragraph{Extending Authentic Triplets w/-LangID (Only Authentic w/-LangID)} This experiment uses multilingual synthetic APE data as used in the \textit{w/o-LangID} experiment. However, we prepend a respective language ID to a source sentence in each multilingual authentic APE dataset triplet.

\paragraph{Extending Synthetic and Authentic Triplets w/-LangID (w/-LangID)} In this experiment, we prepend a respective language ID to a source sentence in each APE triplet of the multilingual synthetic and multilingual authentic APE dataset.

Based on the experimental results, we perform further MAPE exploration considering \textbf{w/-LangID} as the base experiment.

\subsection{Data Augmentation}
Through the following experiment, we try to assess the effectiveness of our proposed data augmentation approaches.

For the data augmentation, we translate a randomly picked subset of 0.5M (based on empirical findings; Refer Appendix \ref{ap:exp_details}) source sentences from the synthetic APE triplets of each pair using IndicTrans2 \cite{gala2023indictrans} NMT system into the `cross-target-language.' We refer to these translations as \textit{External translations}. For example, we translate the source sentences from the English-Hindi synthetic APE data into the Marathi language to get quadruples of the following form: \textit{<English source, Marathi External translation, Hindi translation, Hindi post-edit>}. Similarly, we translate the source sentences from the English-Marathi synthetic APE data into Hindi to get the \textit{<English source, Hindi External translation, Marathi translation, Marathi post-edit>} quadruples.

The data augmentation is performed only during the second stage of CTS, in which the model is trained using the multilingual synthetic APE data. The authentic APE data is not augmented.

\paragraph{w/-LangID + Additional Pairs} In this experiment, we exploit the cross-lingual information to improve the understanding of the source encoder by providing the APE model triplets of additional language pairs. 

We use the quadruples to form Hindi-Marathi (\textit{<Hindi External translation, Marathi translation, Marathi post-edit>}) and Marathi-Hindi (\textit{<Marathi external translation, Hindi translation, Hindi post-edit>}) APE triplets. These triplets are added to the combined English-Hindi and English-Marathi synthetic APE datasets, and the entire set is shuffled before being used for the training. As in the case of the \textit{w/-LangID} experiment, the source sentences in the augmented triplets are also prefixed with their respective language ID tag.

Therefore, during the second stage of the CTS, the model is trained on English-Hindi, English-Marathi, Hindi-Marathi, and Marathi-Hindi APE data.

\paragraph{w/-LangID + External Candidates} We extend the data augmentation approach proposed by \citet{yang-etal-2020-hw} to the MAPE setting. In this experiment, we try to enhance the quality of the translation encoder by additionally providing it with the cross-lingual translation by appending it to the existing translation after adding a \textit{`<sep>'} token. Therefore, the synthetic English-Hindi APE triplets are modified as follows: \textit{<English source, Hindi translation <sep> Marathi external translation, Hindi post-edit>}. Similarly, the synthetic English-Marathi APE triplets take the following form: \textit{<English source, Marathi translation <sep> Hindi external translation, Marathi post-edit>}. 

\subsection{APE-QE Multitask Learning}

The sentence-level and word-level QE tasks have been shown to be helpful in improving the performance of APE systems in a scenario where the model is jointly trained for the QE and APE tasks in a single-pair setting. Through this experiment, we explore whether using sentence-level and word-level QE tasks along with APE also leads to improvements in the MAPE scenario. The following two experiments add to the setting of the \textit{w/-LangID} experiment.

These experiments add multitask-learning to the \textit{w/-LangID} experiment. During the fine-tuning stage of the CTS, the MAPE model is trained on the sentence-level QE and word-level QE tasks as well. We set the sentence-level QE loss to zero for those instances in the authentic APE dataset for which the Direct Assessment annotations are unavailable.

We perform the following two experiments under the multitask learning setting. (1) \textbf{MTL-MAPE (LS-MTL)}: The linear scalarization is used for combining task-specific losses. (2) \textbf{MTL-MAPE (Nash)}: Nash-MTL technique is used to combine the gradients and update the model weights. Refer to \citet{Navon2022MultiTaskLA} or \citet{deoghare-etal-2023-quality} for details on Nash-MTL.

\subsection{Domain Adaptation} In this experiment, which we refer to as \textbf{DomainAdapt}, we explore the possibility of improving the domain-specific performance of MAPE through the domain-specific sentence-level and word-level QE information incorporated through multitask learning. In this experiment, during the fine-tuning stage of the CTS, the model is trained only on the domain-specific APE-QE instances of both language pairs. Similar to the \textit{MTL-MAPE} experiments, we set the sentence-level QE loss to zero for those instances in the authentic APE dataset for which the Direct Assessment annotations are unavailable. Therefore, through this experiment, we get a separate APE model for each domain.

Since we end up with only a few training instances for each domain during fine-tuning, updating all network parameters during fine-tuning results in over-fitting. To avoid it, we add a single adapter layer in each decoder block as in \citet{huang-etal-2022-luls} and only update these adapter layers during fine-tuning. The overhead in performing this experiment is we get a separate domain-specific checkpoint. We combine news- and tourism-domain authentic APE triplets for this experiment for both pairs. Also, we combine the En-Mr health and En-Hi law and tourism domain triplets to form a `General' domain authentic MAPE data.


\section{Datasets}
The MT and APE datasets of English-Hindi and English-Marathi are smaller than other language pairs like English-German, English-Russian, and English-Chinese. Considering the amount of MT parallel corpus, synthetic APE triplets, and authentic APE triplets for these language pairs, we consider English-Hindi and English-Marathi to be low-resource language pairs.

For English-Marathi APE, we use datasets released through the WMT22~\cite{ bhattacharyya-etal-2022-findings} English-Marathi APE shared task. The APE data consists of 18K authentic (human-generated post-edits) and 2.5M synthetic~\cite{negri-etal-2018-escape} APE triplets. The authentic APE triplets come from the News (6.5K), Tourism (6.5K), and Health (5K) domains, and the synthetic APE data triplets from multiple domains.

We use an in-house created APE corpus consisting of 9K training triplets for the English-Hindi pair, as an open-source APE corpus for this language pair is unavailable. Source sentences used in the authentic APE corpus are taken from publicly available data. 2.25K authentic triplets come from each of the News, Tourism, Law, and Education domains. Also, the MT system used for translating the source sentences is open-source. Apart from the training set, additional 1K-1K authentic triplets are set aside to be used as development and test sets. The source sentences in these datasets cover multiple domains.
Additionally, we generate multi-domain 2.5M synthetic APE triplets from a subset of the Anuvaad\footnote{\href{https://github.com/project-anuvaad/anuvaad-parallel-corpus}{Anuvaad Parallel Corpus}} parallel corpus following the approach proposed by \citet{negri-etal-2018-escape}. Along with this corpora, its details, annotation process, and guidelines are planned to be released through the upcoming WMT APE shared task. Further, we also use the BPCC~\cite{gala2023indictrans} corpus for both pairs to train an MT model during the first stage of the CTS.

We use the English-Marathi and English-Hindi sentence-level QE data released through the WMT22 \citet{zerva-etal-2022-findings} and WMT23 \cite{blain-etal-2023-findings} QE DA shared tasks, respectively, to get the DA scores for the APE triplets. We generate the word-level QE tags for both the pairs through the QE-corpus-builder tool\footnote{\url{https://github.com/deep-spin/qe-corpus-builder}}.

\section{Results and Discussion}
This section discusses the results of different experiments. For the quantitative evaluation, we consider TER~\cite{snover-etal-2006-study} as the primary metric. We also report BLEU~\cite{papineni-etal-2002-bleu} scores. We perform a statistical significance test considering primary metric (TER) using William's significance test~\cite{graham-2015-improving}. The model training approach and the training details used are discussed in Appendix \ref{ap:exp_details}.

\begin{table}[]
\resizebox{\columnwidth}{!}{%
\begin{tabular}{l|rrrr}
\hline
\multicolumn{1}{c|}{\textbf{Technique}} & \multicolumn{2}{c}{\textbf{En-Hi}}                                   & \multicolumn{2}{c}{\textbf{En-Mr}}                                   \\
\textbf{}                               & \multicolumn{1}{c}{\textbf{TER}} & \multicolumn{1}{c}{\textbf{BLEU}} & \multicolumn{1}{c}{\textbf{TER}} & \multicolumn{1}{c}{\textbf{BLEU}} \\ \hline
\textbf{Do Nothing}                     & 27.76                            & 58.86                             & 22.93                            & 64.51                             \\
\textbf{Baseline APE}                  & 20.85                            & 66.70                             & 20.58                            & 66.95                             \\
\textbf{Transfer Learning}              & 20.59                            & 66.99                             & 20.25*                            & 67.38                             \\
\textbf{w/o-LangID}                    & 19.46                            & 68.18                             & 19.55                            & 68.11                             \\
\textbf{Only Authentic w/-LangID}  & 18.93                            & 68.91                             & 18.73                            & 69.04                             \\
\textbf{w/-LangID}                     & \textbf{18.35}                   & \textbf{69.56}                    & \textbf{18.19}                   & \textbf{69.56}                    \\ \hline
\end{tabular}%
}
\caption{Results of English-Hindi and English-Marathi APE systems on their respective evaluation sets. \textbf{*} denotes that the improvement over the primary baseline (\textit{Baseline APE}) is insignificant ($p$ being 0.05)}
\label{tab:res1}
\end{table}

\paragraph{Multilingual APE} Table \ref{tab:res1} compiles the results of initial multilingual experiments performed to investigate the impact of multilingual training and the importance of embedding explicit information about the target language into the MAPE model input. The model weight initialization through the other trained APE model (\textit{Transfer Learning}) does not yield robust improvements even in the case of close target languages. The improvements are significant in the case of the En-Hi APE, while they are insignificant in the case of the En-Mr APE. It hints at the En-Mr APE model or data being more helpful to En-Hi APE than the other way around. A possible reason for such an outcome could be comparatively less authentic training data for En-Hi APE and a relatively easy evaluation set.

Comparison between the single-pair APE models and the MAPE model trained without inputting explicit information about the target language of the triplets reveals the potential of multilingual training in the case of closely related target languages. For both pairs, we observe more than 1 TER point improvement. In this comparison, we also observe En-Hi APE achieving more improvements than En-Mr APE. 

The positive impact of presenting explicit information about the target language is evident from the results. Greater improvements when this information is presented to the model for all the triplets (synthetic and authentic) over the improvements when passed only for the authentic APE triplets highlight the need for target language token prefixing to the APE model while training in a multilingual setup.

The MAPE results show impressive improvements of 2.5 and 2.39 TER points over their corresponding single-pair APE models for En-hi and En-Mr, respectively, underlining the effectiveness of multilingualism in the case of linguistically similar languages.

\paragraph{Data Augmentation} Results of data augmentation experiments are depicted in Table \ref{tab:res2}. The results show better performance gains through cross-lingual transfer when the multilingual model is trained on additional post-editing directions. Though not great, we observe modest gains of around 0.5 TER points for both pairs. Unlike the vanilla MAPE experiments, in the case of these experiments, we observe both pairs benefiting equally. 

Comparatively, data augmentation through external candidates does not seem to be of much help as the model shows insignificant improvement in performance for En-Mr and small gains for the En-Hi pair. A possible reason could be that the translation encoder is already exposed to a vast amount of Hindi and Marathi text.

\begin{table}[]
\resizebox{\columnwidth}{!}{%
\begin{tabular}{l|rrrr}
\hline
\multicolumn{1}{c|}{\textbf{}}   & \multicolumn{2}{c}{\textbf{En-Hi}}                                   & \multicolumn{2}{c}{\textbf{En-Mr}}                                   \\
\textbf{}                        & \multicolumn{1}{c}{\textbf{TER}} & \multicolumn{1}{c}{\textbf{BLEU}} & \multicolumn{1}{c}{\textbf{TER}} & \multicolumn{1}{c}{\textbf{BLEU}} \\ \hline
\textbf{Do Nothing}              & 27.76                            & 58.86                             & 22.93                            & 64.51                             \\
\textbf{Baseline APE}           & 20.85                            & 66.70                             & 20.58                            & 66.95                             \\
\textbf{w/-LangID}              & 18.35                            & 69.56                             & 18.19                            & 69.56                             \\
\textbf{w/-LangID + Additional Pairs}    & \textbf{17.82}                   & \textbf{70.10}                    & \textbf{17.74}                   & \textbf{70.23}                    \\
\textbf{w/-LangID + External Candidates} & 17.94                            & 69.82                             & 17.98*                           & 69.77                             \\ \hline
\end{tabular}%
}
\caption{Results of English-Hindi and English-Marathi APE systems on their respective evaluation sets. \textbf{*} denotes the improvement over the \textit{w/ LangID} is insignificant ($p$ being 0.05).}
\label{tab:res2}
\end{table}

\paragraph{Multitask Learning} Table \ref{tab:res3} shows the extent of benefit that APE receives in the multilingual scenario from QE tasks through multitask learning. Similar to how joint training on QE and APE tasks has been shown to benefit APE in the case of single-pair APE models, results show its usefulness in the MAPE setting as well. Even the MAPE model trained through the simplest multitask learning approach, where the task-specific losses are added together, shows significant improvements over the non-multitask learning-based MAPE model. As expected, we observe further performance improvements for both pairs when a more sophisticated approach is employed to perform multitask learning.

When we use the MAPE model trained using the data augmentation through additional translation directions and fine-tune it through multitask learning, we do not see significant improvements over the multitask learning-based MAPE model trained without the data augmentation. This could be due to the law of diminishing returns. However, through this combination, we achieve improvements of more than 3 TER points for both pairs over their single-pair-based APE counterparts.

To the best of our knowledge, 16.75 TER points on the WMT22 English-Marathi APE development set are the highest performance achieved so far.

\begin{table}[]
\resizebox{\columnwidth}{!}{%
\begin{tabular}{l|rrrr}
\hline
\multicolumn{1}{c|}{\textbf{Technique}} & \multicolumn{2}{c}{\textbf{En-Hi}}                                   & \multicolumn{2}{c}{\textbf{En-Mr}}                                   \\
\textbf{}                               & \multicolumn{1}{c}{\textbf{TER}} & \multicolumn{1}{c}{\textbf{BLEU}} & \multicolumn{1}{c}{\textbf{TER}} & \multicolumn{1}{c}{\textbf{BLEU}} \\ \hline
\textbf{Do Nothing}                     & 27.76                            & 58.86                             & 22.93                            & 64.51                             \\
\textbf{Baseline APE}                  & 20.85                            & 66.70                             & 20.58                            & 66.95                             \\
\textbf{w/-LangID}                     & 18.35                            & 69.56                             & 18.19                            & 69.56                             \\
\textbf{MTL-MAPE (LS-MTL))}                       & 17.60                            & 70.37                             & 17.36                            & 70.52                             \\
\textbf{MTL-MAPE (Nash)}                & 17.28                            & 70.75                             & 16.90                            & 71.01                             \\
\textbf{MTL-MAPE (Nash) + DataAug}      & \textbf{17.06}                  & \textbf{70.96}                    & \textbf{16.75}                  & \textbf{71.14}                    \\ \hline
\end{tabular}%
}
\caption{Results of Multitask Learning-based multilingual En-Hi and En-Mr APE experiments. The last row refers to an additional experiment where Nash-MTL-based multitask learning is used during the fine-tuning stage in the \textit{w/-LangID + Additional Pairs} experiment. All improvements are significant with respect to the \textit{w/-LangID} experiment ($p$ being 0.05).}
\label{tab:res3}
\end{table}

\paragraph{Domain Adaptation} Table \ref{tab:res5} shows the result of the domain adaptation experiment. The TER and BLEU scores reported in the table in the \textit{DomainAdapt} row are an average of domain-specific scores obtained from separate domain-specific checkpoints. The domain adaption through multitask learning yields significant but modest performance improvements. Even though the adapters are used during the fine-tuning to prevent overfitting, they do not surpass or achieve comparable performance as we observe for a model, which is trained through data augmentation and fine-tuned through the QE-APE multitask learning.

\begin{table}[]
\resizebox{\columnwidth}{!}{%
\begin{tabular}{l|rrrr}
\hline
\multicolumn{1}{c|}{\textbf{Technique}} & \multicolumn{2}{c}{\textbf{En-Hi}}                                   & \multicolumn{2}{c}{\textbf{En-Mr}}                                   \\
\textbf{}                               & \multicolumn{1}{c}{\textbf{TER}} & \multicolumn{1}{c}{\textbf{BLEU}} & \multicolumn{1}{c}{\textbf{TER}} & \multicolumn{1}{c}{\textbf{BLEU}} \\ \hline
\textbf{Do Nothing}                     & 27.76                            & 58.86                             & 22.93                            & 64.51                             \\
\textbf{Baseline APE}                  & 20.85                            & 66.70                             & 20.58                            & 66.95                             \\
\textbf{w/ LangID}                     & 18.35                            & 69.56                             & 18.19                            & 69.56                             \\
\textbf{DomainAdapt}                    & \textbf{18.00}                   & \textbf{70.01}                    & \textbf{17.74}                   & \textbf{69.98}                    \\ \hline
\end{tabular}%
}
\caption{Results of Multitask Learning-based domain adaptation multilingual En-Hi and En-Mr APE experiments. All improvements with respect to w/ LangID are significant (with p being 0.05).}
\label{tab:res5}
\end{table}

\begin{figure*}[t]
\centering
\includegraphics[width=\linewidth]{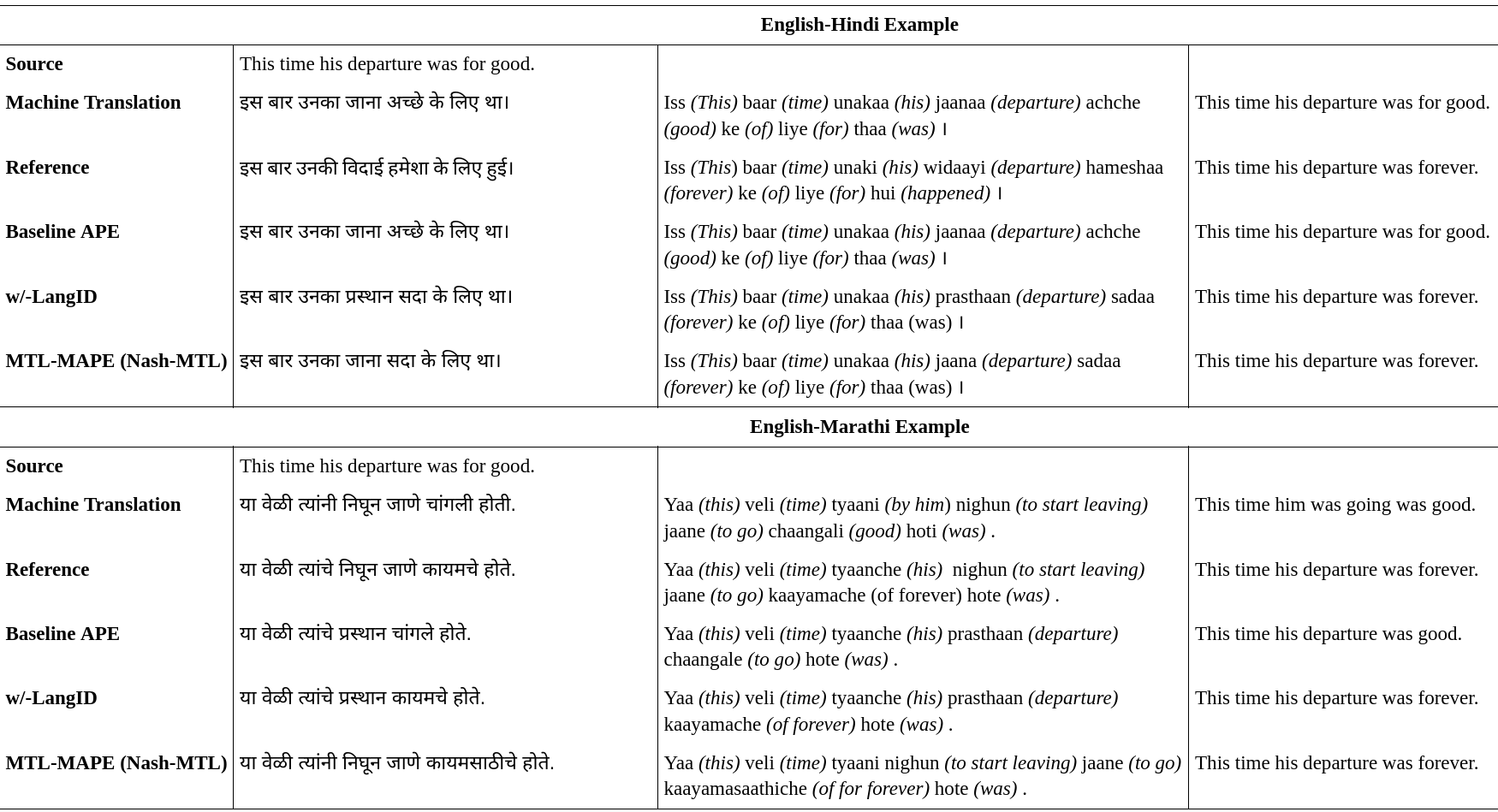}
\caption{Comparison of English-Marathi and English-Hindi APE outputs obtained from Baseline APE and two MAPE systems.}
\label{fig:qual}
\end{figure*}

\paragraph{Case Analysis} Figure \ref{fig:qual} shows two representative English-Hindi and English-Marathi APE examples to see whether the MAPE TER improvements reflect in the model outputs and also to investigate how multilingualism and multilingual QE-APE joint training helps the MAPE model.

The Hindi machine translation of the English source sentence is literal and misses the sense of the entity's departure being forever. The \textit{Baseline APE} does not make any changes to the translation, which shows it failed to recognize the quality of the translation. Impressively, the MAPE successfully understands the literal translation of 'good' is incorrect and correctly uses the Hindi word \textit{`sadaa'} instead of \textit{`achche'} in the post-edit. It suggests how multilingual training improves the overall understanding of the target languages.

However, we see the MAPE model deviating from the \textit{principle of minimal editing} as it unnecessarily changes the Hindi translation of the word `departure' from \textit{`jaanaa'} to its synonym \textit{`prasthaan'}. It suggests the need to take steps to mitigate the problem of over-correction. While \textit{`prasthaan'} word is present in both languages with the same meaning, its usage in Marathi is relatively more. The presence of the word \textit{`prasthaan'} in the English-Marathi \textit{Baseline APE} output again points to the knowledge sharing from English-Marathi to English-Hindi, due to multilingual training.

The MTL-MAPE model output where the translation of `departure' is not touched and the correct translation to convey the sense of `good' shows that training the MAPE model jointly on APE and QE tasks helps the model to follow the `principle of minimal editing.' While the reference post-edit is far from the \textit{MTL-MAPE} output, both sentences can be considered good translations.

On the English-Marathi side, the machine translation of the same source sentence is literal and also not fluent. The English-Marathi \textit{Baseline APE} corrected all the fluency issues in the translation. However, it still is the literal translation of the source. The \textit{MAPE} model takes care of the literal translation issue by making the necessary changes. As observed in the case of English-Hindi output of the \textit{MAPE}, we observe the use of \textit{`prasthaan'} instead of using the same Marathi phrase \textit{`nighun jaane'} in the Marathi post-edit as well. Further, we see this issue being taken care of by the \textit{MTL-MAPE} model, as also seen in the case of the Hindi post-editing. 

Such similarity in the behavior of the \textit{MAPE} and \textit{MTL-MAPE} models suggest that the MAPE training might be helping the models in generating consistent post-edits across the target languages when the source sentence is the same, and the quality of the machine translations is similar. While not reported in the paper, we would like to share that our analysis of a small subset of instances and their corresponding MAPE outputs strengthens this conjecture. However, a large-scale scientific analysis is required to confirm this claim. We provide additional examples in Appendix \ref{ap:add_eg}.

\section{Conclusion and Future Work} 

In this work, we explore multilingual APE, which has not yet been attempted to the best of our knowledge. We focus on two low-resource English-Indic language pairs where the target languages, Hindi and Marathi, are closely related Indo-Aryan languages. The results of our initial experiments show the effectiveness of multilingual training for improving APE performance of both pairs. Relatively higher improvements for English-Hindi than English-Marathi suggest that the MAPE can benefit the lower resource language pair in the APE setting, to improve upon their APE performance. 


We observe further performance gains using the proposed simple yet effective data augmentation approach in which we generate synthetic Hindi-Marathi and Marathi-Hindi APE triplets, to enable cross-lingual transfer. Our investigation into the use of the QE-APE multitask learning framework in the multilingual APE setting reveals that the QE helps MAPE improve its capability to assess translation quality, and mitigate the problem of over-correction. The exploration of using only the domain-specific QE information to perform domain adaptation shows that modest performance gains can also be obtained, even in resource-constrained settings.

With $17.06$ and $16.75$ TER on English-Hindi and English-Marathi evaluation sets, our multilingual APE model achieves state-of-the-art performance for the English-Marathi pair, and establishes a strong baseline for English-Hindi. Also, our qualitative analysis hints at the possible future work- Using MAPE for consistent post-edits across languages.

The investigation performed in this work suggests that multilingual APE is a promising research direction for the advancement of APE systems for low-resource language pairs. Even though the gains of more than 3 TER points highlight the robustness of MAPE, in the future, this work can be extended to multiple low-resource language pairs to analyze the generalizability. An in-depth study could be conducted to analyze the extent of consistency of the MAPE model outputs across target languages when fed with the same source and similar quality translations. Further experiments could be conducted to gauge the optimal number of augmented triplets of specific additional language pairs to improve the MAPE performance on specific or all language pairs. 

\section{Limitations}
The multilingual APE investigation undertaken under this work is limited to a specific case of two low-resource language pairs where the source language is English, and the target language is either Hindi or Marathi, which are linguistically closely related Indian languages. Due to the lack of APE resources for other language pairs, answering whether the simple multilingual APE training, as explored in this work, will result in impressive performance gains for other low-resource language pairs as well needs further exploration. Also, the quality of \textit{Baseline APE} models of both pairs (20.85 and 20.58 TER points) is similar. Current work does not explore the improvements MAPE can bring when the \textit{Baseline APE} models of each pair have different qualities.

\section{Ethics Statement}
Our APE models are trained on either in-house datasets or on the publicly available datasets referenced in this paper. These datasets have been previously collected and annotated; no new data collection has been carried out as part of this work. The in-house created dataset will be made public through the upcoming WMT APE shared task. Furthermore, these are standard benchmarks released in recent WMT shared tasks. No user information was present in any of the datasets used in the work, protecting the privacy and identity of users. Also, the synthetic data, code, and models generated as a part of this work will be released publicly under the CC-BY-SA 4.0 license for further research. We understand that every dataset is subject to intrinsic bias and that computational models will inevitably learn biased information from any dataset.

\bibliography{anthology,custom}

\begin{thebibliography}{37}
\expandafter\ifx\csname natexlab\endcsname\relax\def\natexlab#1{#1}\fi

\bibitem[{Aharoni et~al.(2019)Aharoni, Johnson, and Firat}]{aharoni-etal-2019-massively}
Roee Aharoni, Melvin Johnson, and Orhan Firat. 2019.
\newblock \href {https://doi.org/10.18653/v1/N19-1388} {Massively multilingual neural machine translation}.
\newblock In \emph{Proceedings of the 2019 Conference of the North {A}merican Chapter of the Association for Computational Linguistics: Human Language Technologies, Volume 1 (Long and Short Papers)}, pages 3874--3884, Minneapolis, Minnesota. Association for Computational Linguistics.

\bibitem[{Akhbardeh et~al.(2021)Akhbardeh, Arkhangorodsky, Biesialska, Bojar, Chatterjee, Chaudhary, Costa-jussa, Espa{\~n}a-Bonet, Fan, Federmann, Freitag, Graham, Grundkiewicz, Haddow, Harter, Heafield, Homan, Huck, Amponsah-Kaakyire, Kasai, Khashabi, Knight, Kocmi, Koehn, Lourie, Monz, Morishita, Nagata, Nagesh, Nakazawa, Negri, Pal, Tapo, Turchi, Vydrin, and Zampieri}]{akhbardeh-etal-2021-findings}
Farhad Akhbardeh, Arkady Arkhangorodsky, Magdalena Biesialska, Ond{\v{r}}ej Bojar, Rajen Chatterjee, Vishrav Chaudhary, Marta~R. Costa-jussa, Cristina Espa{\~n}a-Bonet, Angela Fan, Christian Federmann, Markus Freitag, Yvette Graham, Roman Grundkiewicz, Barry Haddow, Leonie Harter, Kenneth Heafield, Christopher Homan, Matthias Huck, Kwabena Amponsah-Kaakyire, Jungo Kasai, Daniel Khashabi, Kevin Knight, Tom Kocmi, Philipp Koehn, Nicholas Lourie, Christof Monz, Makoto Morishita, Masaaki Nagata, Ajay Nagesh, Toshiaki Nakazawa, Matteo Negri, Santanu Pal, Allahsera~Auguste Tapo, Marco Turchi, Valentin Vydrin, and Marcos Zampieri. 2021.
\newblock \href {https://aclanthology.org/2021.wmt-1.1} {Findings of the 2021 conference on machine translation ({WMT}21)}.
\newblock In \emph{Proceedings of the Sixth Conference on Machine Translation}, pages 1--88, Online. Association for Computational Linguistics.

\bibitem[{Bhattacharyya et~al.(2022)Bhattacharyya, Chatterjee, Freitag, Kanojia, Negri, and Turchi}]{bhattacharyya-etal-2022-findings}
Pushpak Bhattacharyya, Rajen Chatterjee, Markus Freitag, Diptesh Kanojia, Matteo Negri, and Marco Turchi. 2022.
\newblock \href {https://aclanthology.org/2022.wmt-1.5} {Findings of the {WMT} 2022 shared task on automatic post-editing}.
\newblock In \emph{Proceedings of the Seventh Conference on Machine Translation (WMT)}, pages 109--117, Abu Dhabi, United Arab Emirates (Hybrid). Association for Computational Linguistics.

\bibitem[{Bhattacharyya et~al.(2023)Bhattacharyya, Chatterjee, Freitag, Kanojia, Negri, and Turchi}]{bhattacharyya-etal-2023-findings}
Pushpak Bhattacharyya, Rajen Chatterjee, Markus Freitag, Diptesh Kanojia, Matteo Negri, and Marco Turchi. 2023.
\newblock \href {https://doi.org/10.18653/v1/2023.wmt-1.55} {Findings of the {WMT} 2023 shared task on automatic post-editing}.
\newblock In \emph{Proceedings of the Eighth Conference on Machine Translation}, pages 672--681, Singapore. Association for Computational Linguistics.

\bibitem[{Blain et~al.(2023)Blain, Zerva, Rei, Guerreiro, Kanojia, C.~de Souza, Silva, Vaz, Jingxuan, Azadi, Orasan, and Martins}]{blain-etal-2023-findings}
Frederic Blain, Chrysoula Zerva, Ricardo Rei, Nuno~M. Guerreiro, Diptesh Kanojia, Jos{\'e}~G. C.~de Souza, Beatriz Silva, T{\^a}nia Vaz, Yan Jingxuan, Fatemeh Azadi, Constantin Orasan, and Andr{\'e} Martins. 2023.
\newblock \href {https://doi.org/10.18653/v1/2023.wmt-1.52} {Findings of the {WMT} 2023 shared task on quality estimation}.
\newblock In \emph{Proceedings of the Eighth Conference on Machine Translation}, pages 629--653, Singapore. Association for Computational Linguistics.

\bibitem[{Bojar et~al.(2016)Bojar, Chatterjee, Federmann, Graham, Haddow, Huck, Jimeno~Yepes, Koehn, Logacheva, Monz, Negri, N{\'e}v{\'e}ol, Neves, Popel, Post, Rubino, Scarton, Specia, Turchi, Verspoor, and Zampieri}]{bojar-etal-2016-findings}
Ond{\v{r}}ej Bojar, Rajen Chatterjee, Christian Federmann, Yvette Graham, Barry Haddow, Matthias Huck, Antonio Jimeno~Yepes, Philipp Koehn, Varvara Logacheva, Christof Monz, Matteo Negri, Aur{\'e}lie N{\'e}v{\'e}ol, Mariana Neves, Martin Popel, Matt Post, Raphael Rubino, Carolina Scarton, Lucia Specia, Marco Turchi, Karin Verspoor, and Marcos Zampieri. 2016.
\newblock \href {https://doi.org/10.18653/v1/W16-2301} {Findings of the 2016 conference on machine translation}.
\newblock In \emph{Proceedings of the First Conference on Machine Translation: Volume 2, Shared Task Papers}, pages 131--198, Berlin, Germany. Association for Computational Linguistics.

\bibitem[{Bojar et~al.(2015)Bojar, Chatterjee, Federmann, Haddow, Huck, Hokamp, Koehn, Logacheva, Monz, Negri, Post, Scarton, Specia, and Turchi}]{bojar-etal-2015-findings}
Ond{\v{r}}ej Bojar, Rajen Chatterjee, Christian Federmann, Barry Haddow, Matthias Huck, Chris Hokamp, Philipp Koehn, Varvara Logacheva, Christof Monz, Matteo Negri, Matt Post, Carolina Scarton, Lucia Specia, and Marco Turchi. 2015.
\newblock \href {https://doi.org/10.18653/v1/W15-3001} {Findings of the 2015 workshop on statistical machine translation}.
\newblock In \emph{Proceedings of the Tenth Workshop on Statistical Machine Translation}, pages 1--46, Lisbon, Portugal. Association for Computational Linguistics.

\bibitem[{Chatterjee et~al.(2020)Chatterjee, Freitag, Negri, and Turchi}]{chatterjee-etal-2020-findings}
Rajen Chatterjee, Markus Freitag, Matteo Negri, and Marco Turchi. 2020.
\newblock \href {https://aclanthology.org/2020.wmt-1.75} {Findings of the {WMT} 2020 shared task on automatic post-editing}.
\newblock In \emph{Proceedings of the Fifth Conference on Machine Translation}, pages 646--659, Online. Association for Computational Linguistics.

\bibitem[{Chatterjee et~al.(2018{\natexlab{a}})Chatterjee, Negri, Rubino, and Turchi}]{chatterjee-etal-2018-findings}
Rajen Chatterjee, Matteo Negri, Raphael Rubino, and Marco Turchi. 2018{\natexlab{a}}.
\newblock \href {https://doi.org/10.18653/v1/W18-6452} {Findings of the {WMT} 2018 shared task on automatic post-editing}.
\newblock In \emph{Proceedings of the Third Conference on Machine Translation: Shared Task Papers}, pages 710--725, Belgium, Brussels. Association for Computational Linguistics.

\bibitem[{Chatterjee et~al.(2018{\natexlab{b}})Chatterjee, Negri, Turchi, Blain, and Specia}]{chatterjee-etal-2018-combining}
Rajen Chatterjee, Matteo Negri, Marco Turchi, Fr{\'e}d{\'e}ric Blain, and Lucia Specia. 2018{\natexlab{b}}.
\newblock \href {https://aclanthology.org/W18-1804} {Combining quality estimation and automatic post-editing to enhance machine translation output}.
\newblock In \emph{Proceedings of the 13th Conference of the Association for Machine Translation in the {A}mericas (Volume 1: Research Track)}, pages 26--38, Boston, MA. Association for Machine Translation in the Americas.

\bibitem[{Chatterji(1969)}]{chatterji1969indo}
S.K. Chatterji. 1969.
\newblock \href {https://books.google.co.in/books?id=639jAAAAMAAJ} {\emph{Indo-Aryan \& Hindi}}.
\newblock Firma K. L. Mukhopadhyay.

\bibitem[{Costa-juss{\`a} et~al.(2022)Costa-juss{\`a}, Cross, {\c{C}}elebi, Elbayad, Heafield, Heffernan, Kalbassi, Lam, Licht, Maillard et~al.}]{costa2022no}
Marta~R Costa-juss{\`a}, James Cross, Onur {\c{C}}elebi, Maha Elbayad, Kenneth Heafield, Kevin Heffernan, Elahe Kalbassi, Janice Lam, Daniel Licht, Jean Maillard, et~al. 2022.
\newblock No language left behind: Scaling human-centered machine translation.
\newblock \emph{arXiv preprint arXiv:2207.04672}.

\bibitem[{Dabre et~al.(2020)Dabre, Chu, and Kunchukuttan}]{dabre2020survey}
Raj Dabre, Chenhui Chu, and Anoop Kunchukuttan. 2020.
\newblock A survey of multilingual neural machine translation.
\newblock \emph{ACM Computing Surveys (CSUR)}, 53(5):1--38.

\bibitem[{Deoghare and Bhattacharyya(2022)}]{deoghare-bhattacharyya-2022-iit}
Sourabh Deoghare and Pushpak Bhattacharyya. 2022.
\newblock \href {https://aclanthology.org/2022.wmt-1.67} {{IIT} {B}ombay{'}s {WMT}22 automatic post-editing shared task submission}.
\newblock In \emph{Proceedings of the Seventh Conference on Machine Translation (WMT)}, pages 682--688, Abu Dhabi, United Arab Emirates (Hybrid). Association for Computational Linguistics.

\bibitem[{Deoghare et~al.(2023)Deoghare, Kanojia, Blain, Ranasinghe, and Bhattacharyya}]{deoghare-etal-2023-quality}
Sourabh Deoghare, Diptesh Kanojia, Fred Blain, Tharindu Ranasinghe, and Pushpak Bhattacharyya. 2023.
\newblock \href {https://doi.org/10.18653/v1/2023.findings-emnlp.115} {Quality estimation-assisted automatic post-editing}.
\newblock In \emph{Findings of the Association for Computational Linguistics: EMNLP 2023}, pages 1686--1698, Singapore. Association for Computational Linguistics.

\bibitem[{Gala et~al.(2023)Gala, Chitale, Raghavan, Gumma, Doddapaneni, M, Nawale, Sujatha, Puduppully, Raghavan, Kumar, Khapra, Dabre, and Kunchukuttan}]{gala2023indictrans}
Jay Gala, Pranjal~A Chitale, A~K Raghavan, Varun Gumma, Sumanth Doddapaneni, Aswanth~Kumar M, Janki~Atul Nawale, Anupama Sujatha, Ratish Puduppully, Vivek Raghavan, Pratyush Kumar, Mitesh~M Khapra, Raj Dabre, and Anoop Kunchukuttan. 2023.
\newblock \href {https://openreview.net/forum?id=vfT4YuzAYA} {Indictrans2: Towards high-quality and accessible machine translation models for all 22 scheduled indian languages}.
\newblock \emph{Transactions on Machine Learning Research}.

\bibitem[{Graham(2015)}]{graham-2015-improving}
Yvette Graham. 2015.
\newblock \href {https://doi.org/10.3115/v1/P15-1174} {Improving evaluation of machine translation quality estimation}.
\newblock In \emph{Proceedings of the 53rd Annual Meeting of the Association for Computational Linguistics and the 7th International Joint Conference on Natural Language Processing (Volume 1: Long Papers)}, pages 1804--1813, Beijing, China. Association for Computational Linguistics.

\bibitem[{Hokamp(2017)}]{hokamp-2017-ensembling}
Chris Hokamp. 2017.
\newblock \href {https://doi.org/10.18653/v1/W17-4775} {Ensembling factored neural machine translation models for automatic post-editing and quality estimation}.
\newblock In \emph{Proceedings of the Second Conference on Machine Translation}, pages 647--654, Copenhagen, Denmark. Association for Computational Linguistics.

\bibitem[{Huang et~al.(2022)Huang, Lou, Zhang, and Mei}]{huang-etal-2022-luls}
Xiaoying Huang, Xingrui Lou, Fan Zhang, and Tu~Mei. 2022.
\newblock \href {https://aclanthology.org/2022.wmt-1.68} {{LUL}{'}s {WMT}22 automatic post-editing shared task submission}.
\newblock In \emph{Proceedings of the Seventh Conference on Machine Translation (WMT)}, pages 689--693, Abu Dhabi, United Arab Emirates (Hybrid). Association for Computational Linguistics.

\bibitem[{Johnson et~al.(2017)Johnson, Schuster, Le, Krikun, Wu, Chen, Thorat, Vi{\'e}gas, Wattenberg, Corrado et~al.}]{johnson2017google}
Melvin Johnson, Mike Schuster, Quoc~V Le, Maxim Krikun, Yonghui Wu, Zhifeng Chen, Nikhil Thorat, Fernanda Vi{\'e}gas, Martin Wattenberg, Greg Corrado, et~al. 2017.
\newblock Google’s multilingual neural machine translation system: Enabling zero-shot translation.
\newblock \emph{Transactions of the Association for Computational Linguistics}, 5:339--351.

\bibitem[{Kakwani et~al.(2020)Kakwani, Kunchukuttan, Golla, N.C., Bhattacharyya, Khapra, and Kumar}]{kakwani-etal-2020-indicnlpsuite}
Divyanshu Kakwani, Anoop Kunchukuttan, Satish Golla, Gokul N.C., Avik Bhattacharyya, Mitesh~M. Khapra, and Pratyush Kumar. 2020.
\newblock \href {https://doi.org/10.18653/v1/2020.findings-emnlp.445} {{I}ndic{NLPS}uite: Monolingual corpora, evaluation benchmarks and pre-trained multilingual language models for {I}ndian languages}.
\newblock In \emph{Findings of the Association for Computational Linguistics: EMNLP 2020}, pages 4948--4961, Online. Association for Computational Linguistics.

\bibitem[{Kanojia et~al.(2020{\natexlab{a}})Kanojia, Dabre, Dewangan, Bhattacharyya, Haffari, and Kulkarni}]{kanojia-etal-2020-harnessing}
Diptesh Kanojia, Raj Dabre, Shubham Dewangan, Pushpak Bhattacharyya, Gholamreza Haffari, and Malhar Kulkarni. 2020{\natexlab{a}}.
\newblock \href {https://doi.org/10.18653/v1/2020.coling-main.119} {Harnessing cross-lingual features to improve cognate detection for low-resource languages}.
\newblock In \emph{Proceedings of the 28th International Conference on Computational Linguistics}, pages 1384--1395, Barcelona, Spain (Online). International Committee on Computational Linguistics.

\bibitem[{Kanojia et~al.(2020{\natexlab{b}})Kanojia, Kulkarni, Bhattacharyya, and Haffari}]{kanojia-etal-2020-challenge}
Diptesh Kanojia, Malhar Kulkarni, Pushpak Bhattacharyya, and Gholamreza Haffari. 2020{\natexlab{b}}.
\newblock \href {https://aclanthology.org/2020.lrec-1.378} {Challenge dataset of cognates and false friend pairs from {I}ndian languages}.
\newblock In \emph{Proceedings of the Twelfth Language Resources and Evaluation Conference}, pages 3096--3102, Marseille, France. European Language Resources Association.

\bibitem[{Lee et~al.(2020)Lee, Lee, Shin, Jung, Kim, and Lee}]{lee-etal-2020-postech}
Jihyung Lee, WonKee Lee, Jaehun Shin, Baikjin Jung, Young-Kil Kim, and Jong-Hyeok Lee. 2020.
\newblock \href {https://aclanthology.org/2020.wmt-1.82} {{POSTECH}-{ETRI}{'}s submission to the {WMT}2020 {APE} shared task: Automatic post-editing with cross-lingual language model}.
\newblock In \emph{Proceedings of the Fifth Conference on Machine Translation}, pages 777--782, Online. Association for Computational Linguistics.

\bibitem[{Lopes et~al.(2019)Lopes, Farajian, Correia, Tr{\'e}nous, and Martins}]{lopes-etal-2019-unbabels}
Ant{\'o}nio~V. Lopes, M.~Amin Farajian, Gon{\c{c}}alo~M. Correia, Jonay Tr{\'e}nous, and Andr{\'e} F.~T. Martins. 2019.
\newblock \href {https://doi.org/10.18653/v1/W19-5413} {Unbabel{'}s submission to the {WMT}2019 {APE} shared task: {BERT}-based encoder-decoder for automatic post-editing}.
\newblock In \emph{Proceedings of the Fourth Conference on Machine Translation (Volume 3: Shared Task Papers, Day 2)}, pages 118--123, Florence, Italy. Association for Computational Linguistics.

\bibitem[{Martins et~al.(2017)Martins, Junczys-Dowmunt, Kepler, Astudillo, Hokamp, and Grundkiewicz}]{martins-etal-2017-pushing}
Andr{\'e} F.~T. Martins, Marcin Junczys-Dowmunt, Fabio~N. Kepler, Ram{\'o}n Astudillo, Chris Hokamp, and Roman Grundkiewicz. 2017.
\newblock \href {https://doi.org/10.1162/tacl_a_00056} {Pushing the limits of translation quality estimation}.
\newblock \emph{Transactions of the Association for Computational Linguistics}, 5:205--218.

\bibitem[{Masica(1993)}]{masica1993indo}
Colin~P Masica. 1993.
\newblock \emph{The indo-aryan languages}.
\newblock Cambridge University Press.

\bibitem[{Navon et~al.(2022)Navon, Shamsian, Achituve, Maron, Kawaguchi, Chechik, and Fetaya}]{Navon2022MultiTaskLA}
Aviv Navon, Aviv Shamsian, Idan Achituve, Haggai Maron, Kenji Kawaguchi, Gal Chechik, and Ethan Fetaya. 2022.
\newblock Multi-task learning as a bargaining game.
\newblock In \emph{International Conference on Machine Learning}.

\bibitem[{Negri et~al.(2018)Negri, Turchi, Chatterjee, and Bertoldi}]{negri-etal-2018-escape}
Matteo Negri, Marco Turchi, Rajen Chatterjee, and Nicola Bertoldi. 2018.
\newblock \href {https://aclanthology.org/L18-1004} {{ESCAPE}: a large-scale synthetic corpus for automatic post-editing}.
\newblock In \emph{Proceedings of the Eleventh International Conference on Language Resources and Evaluation ({LREC} 2018)}, Miyazaki, Japan. European Language Resources Association (ELRA).

\bibitem[{Oh et~al.(2021)Oh, Jang, Xu, An, and Oh}]{oh-etal-2021-netmarble}
Shinhyeok Oh, Sion Jang, Hu~Xu, Shounan An, and Insoo Oh. 2021.
\newblock \href {https://aclanthology.org/2021.wmt-1.34} {Netmarble {AI} center{'}s {WMT}21 automatic post-editing shared task submission}.
\newblock In \emph{Proceedings of the Sixth Conference on Machine Translation}, pages 307--314, Online. Association for Computational Linguistics.

\bibitem[{Papineni et~al.(2002)Papineni, Roukos, Ward, and Zhu}]{papineni-etal-2002-bleu}
Kishore Papineni, Salim Roukos, Todd Ward, and Wei-Jing Zhu. 2002.
\newblock \href {https://doi.org/10.3115/1073083.1073135} {{B}leu: a method for automatic evaluation of machine translation}.
\newblock In \emph{Proceedings of the 40th Annual Meeting of the Association for Computational Linguistics}, pages 311--318, Philadelphia, Pennsylvania, USA. Association for Computational Linguistics.

\bibitem[{Sharma et~al.(2021)Sharma, Gupta, and Nelakanti}]{sharma-etal-2021-adapting}
Abhishek Sharma, Prabhakar Gupta, and Anil Nelakanti. 2021.
\newblock \href {https://aclanthology.org/2021.wmt-1.35} {Adapting neural machine translation for automatic post-editing}.
\newblock In \emph{Proceedings of the Sixth Conference on Machine Translation}, pages 315--319, Online. Association for Computational Linguistics.

\bibitem[{Snover et~al.(2006)Snover, Dorr, Schwartz, Micciulla, and Makhoul}]{snover-etal-2006-study}
Matthew Snover, Bonnie Dorr, Rich Schwartz, Linnea Micciulla, and John Makhoul. 2006.
\newblock \href {https://aclanthology.org/2006.amta-papers.25} {A study of translation edit rate with targeted human annotation}.
\newblock In \emph{Proceedings of the 7th Conference of the Association for Machine Translation in the Americas: Technical Papers}, pages 223--231, Cambridge, Massachusetts, USA. Association for Machine Translation in the Americas.

\bibitem[{Subb{\=a}r{\=a}o(2012)}]{subbarao2012south}
K{\=a}rum{\=u}ri~V Subb{\=a}r{\=a}o. 2012.
\newblock \emph{South Asian languages: A syntactic typology}.
\newblock Cambridge University Press.

\bibitem[{Wei et~al.(2020)Wei, Shang, Wu, Yu, Li, Guo, Wang, Yang, Lei, Qin, and Sun}]{wei-etal-2020-hw}
Daimeng Wei, Hengchao Shang, Zhanglin Wu, Zhengzhe Yu, Liangyou Li, Jiaxin Guo, Minghan Wang, Hao Yang, Lizhi Lei, Ying Qin, and Shiliang Sun. 2020.
\newblock \href {https://aclanthology.org/2020.wmt-1.31} {{HW}-{TSC}{'}s participation in the {WMT} 2020 news translation shared task}.
\newblock In \emph{Proceedings of the Fifth Conference on Machine Translation}, pages 293--299, Online. Association for Computational Linguistics.

\bibitem[{Yang et~al.(2020)Yang, Wang, Wei, Shang, Guo, Li, Lei, Qin, Tao, Sun, and Chen}]{yang-etal-2020-hw}
Hao Yang, Minghan Wang, Daimeng Wei, Hengchao Shang, Jiaxin Guo, Zongyao Li, Lizhi Lei, Ying Qin, Shimin Tao, Shiliang Sun, and Yimeng Chen. 2020.
\newblock \href {https://aclanthology.org/2020.wmt-1.85} {{HW}-{TSC}{'}s participation at {WMT} 2020 automatic post editing shared task}.
\newblock In \emph{Proceedings of the Fifth Conference on Machine Translation}, pages 797--802, Online. Association for Computational Linguistics.

\bibitem[{Zerva et~al.(2022)Zerva, Blain, Rei, Lertvittayakumjorn, C.~de Souza, Eger, Kanojia, Alves, Or{\u{a}}san, Fomicheva, Martins, and Specia}]{zerva-etal-2022-findings}
Chrysoula Zerva, Fr{\'e}d{\'e}ric Blain, Ricardo Rei, Piyawat Lertvittayakumjorn, Jos{\'e}~G. C.~de Souza, Steffen Eger, Diptesh Kanojia, Duarte Alves, Constantin Or{\u{a}}san, Marina Fomicheva, Andr{\'e} F.~T. Martins, and Lucia Specia. 2022.
\newblock \href {https://aclanthology.org/2022.wmt-1.3} {Findings of the {WMT} 2022 shared task on quality estimation}.
\newblock In \emph{Proceedings of the Seventh Conference on Machine Translation (WMT)}, pages 69--99, Abu Dhabi, United Arab Emirates (Hybrid). Association for Computational Linguistics.

\end{thebibliography}

\appendix

\section{Experimental Details}\label{ap:exp_details}
The experimental setup for conducting all APE experiments across English-Marathi (En-Mr) and English-Hindi (En-Hi) language pairs.

\subsection{Training Approach}
We utilize the Curriculum Training Strategy (CTS) as described by \citet{deoghare-etal-2023-quality} to train both APE and MAPE systems. The steps of CTS are outlined as follows:

Initially, we train a single-encoder, single-decoder model for the NMT task. We merge the parallel corpora of both language pairs to train a multilingual NMT. Each source sentence in the corpus is prefixed with the respective target Language ID (`hin\_Deva' for Hindi, `mar\_Deva' for Marathi).

In the next step, we introduce the translation encoder and train the two-encoder single-decoder model for the APE task using synthetic APE data in two phases. In the first phase, the model is trained on a subset of the synthetic corpus consisting of triplets with poorer TER than the Do Nothing baseline. In the second phase, training continues on the remaining synthetic corpus subset, which includes triplets with equal or better TER than the Do Nothing baseline.

Finally, in the third step, we fine-tune the APE model using authentic APE data.

\begin{table}[t!]
\centering
\begin{tabular}{l|rr}
\hline
\multicolumn{1}{c|}{\textbf{No. of Triplets}} & \multicolumn{1}{c}{\textbf{En-Hi}} & \multicolumn{1}{c}{\textbf{En-Mr}} \\ \hline
\textbf{0.1M}                                 & 18.16                              & 18.09                              \\
\textbf{0.25M}                                & 17.95                              & 17.78                              \\
\textbf{0.5M}                                 & \textbf{17.82}                     & \textbf{17.74}                     \\
\textbf{1M}                                   & 17.89                              & 17.81                              \\
\textbf{2.5M}                                 & 18.50                              & 18.30                              \\ \hline
\end{tabular}
\caption{TER scores of \textit{w/-LangID + Additional Pairs} experiment on the respective evaluation set when different amounts of triplets are used for the data augmentation. An equal number of triplets Hindi-Marathi and Marathi-Hindi triplets are used in each experiment.}
\label{tab:data_aug_analysis}
\end{table}

\subsection{Training Details}
We adhere to a standardized set of configurations to ensure uniformity in experimental settings across different experiments. The batch size is 32. We employ early stopping with a patience of 5 epochs and train the model for a maximum of 10000 epochs. The Adam optimizer is used with a learning rate of 5 x \(10^{-5}\), and \(\beta_1\) and \(\beta_2\) are to 0.9 and 0.997, respectively. We incorporate 15,000 warmup steps. For the beam search, the beam size is set to 5. The size of the adapter used in the domain adaptation experiments is 512. NVIDIA100 GPUs are utilized for experimentation purposes. The MAPE model consists of approximately 40 million parameters, and training with CTS takes around 56 hours. Given the extensive number of experiments and the significant time and resources required, we report single-run results.

For preprocessing the data, we utilize the NLTK library\footnote{\url{https://www.nltk.org/}} for English text and the IndicNLP library\footnote{\url{https://github.com/anoopkunchukuttan/indic_nlp_library}} for Hindi and Marathi text. Model training and inference are carried out using Pytorch\footnote{\url{https://pytorch.org/}}. To compute TER scores, we use the official WMT APE evaluation script\footnote{\url{https://github.com/sheffieldnlp/qe-eval-scripts}}, and for BLEU scores, we use the SacreBLEU\footnote{\url{https://github.com/mjpost/sacrebleu}} library.

\begin{figure*}[t]
\centering
\includegraphics[width=\linewidth]{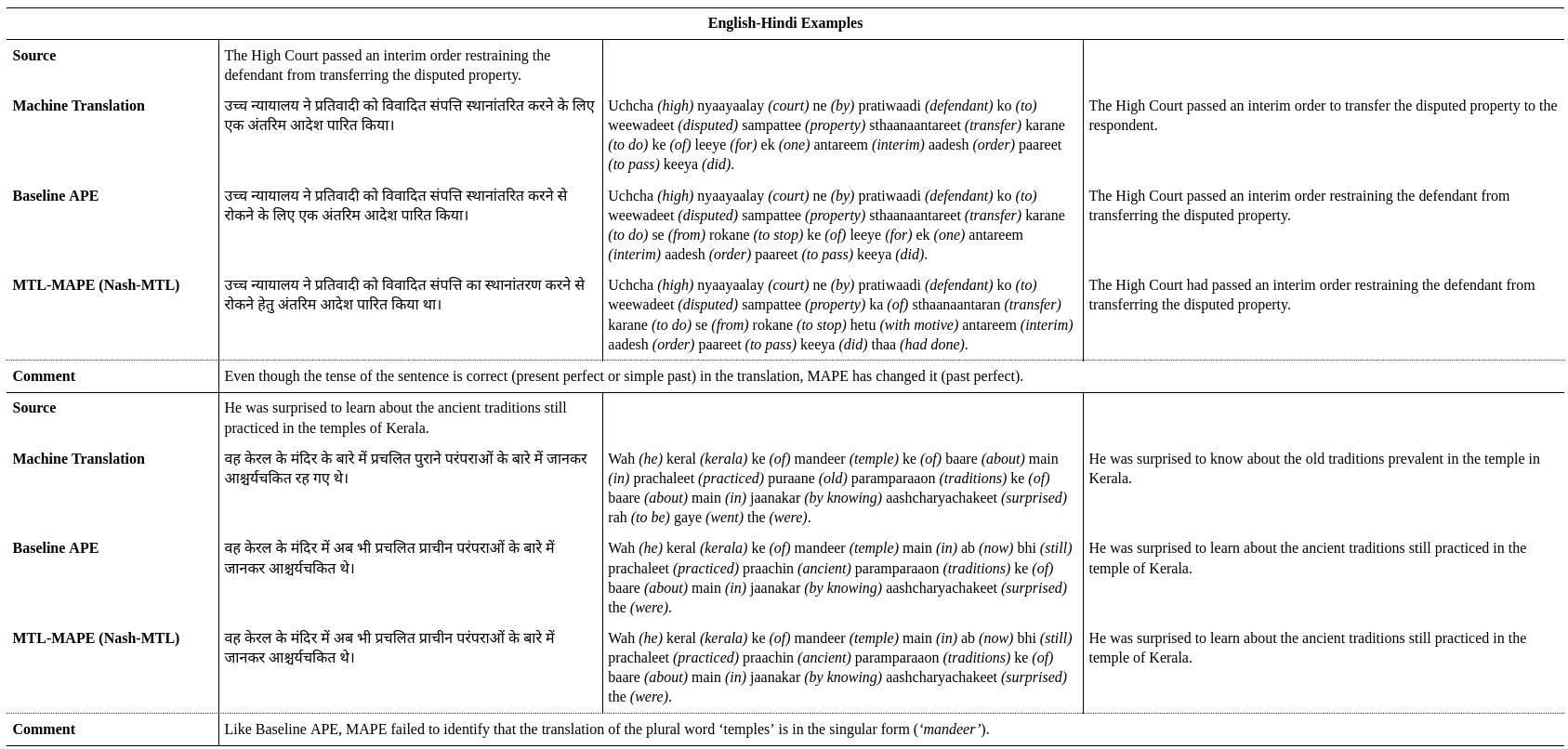}
\caption{Comparison of English-Marathi and English-Hindi APE outputs obtained from Baseline APE and MAPE systems.}
\label{fig:add_examples_en_hi}
\end{figure*}

\begin{figure*}[t]
\centering
\includegraphics[width=\linewidth]{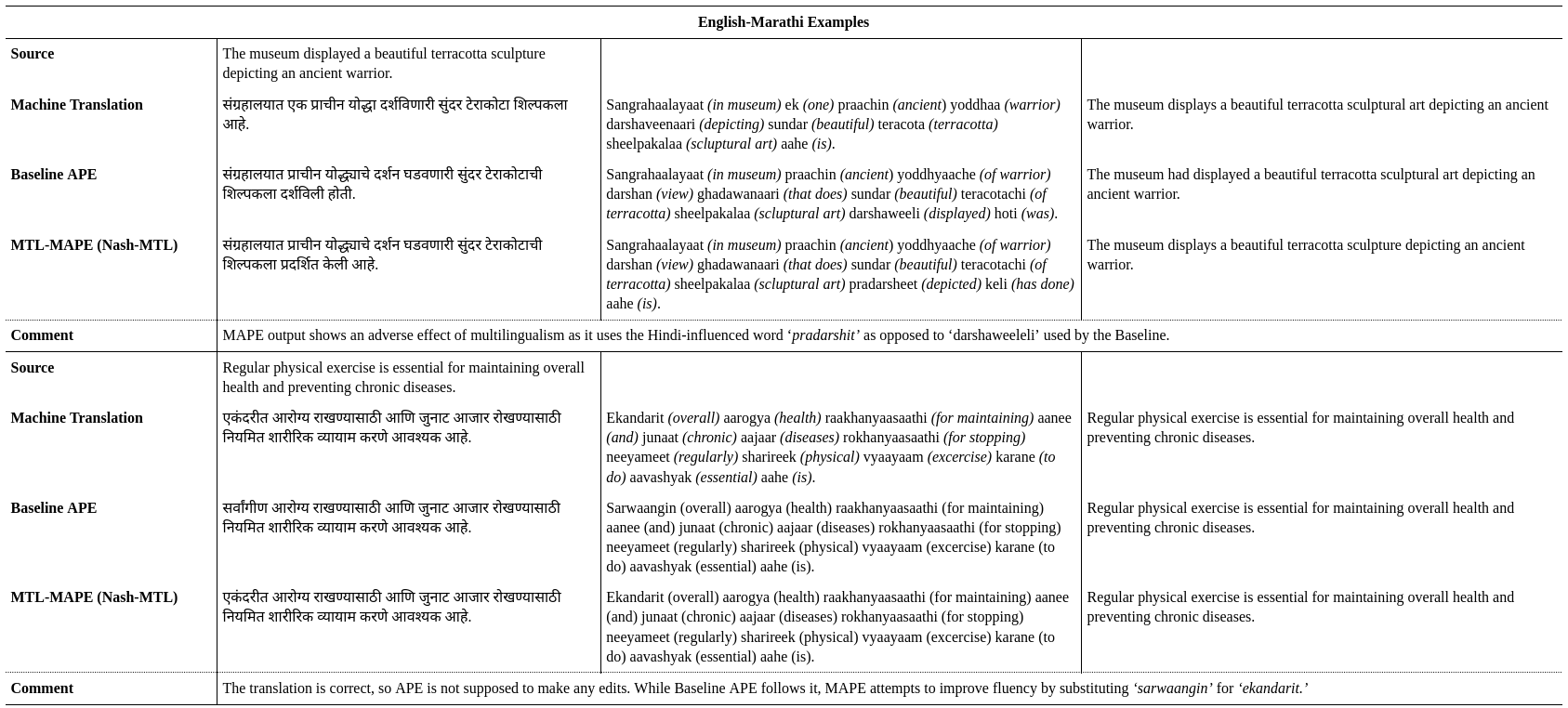}
\caption{Comparison of English-Marathi and English-Hindi APE outputs obtained from Baseline APE and MAPE systems.}
\label{fig:add_examples_en_mr}
\end{figure*}

\section{Data Augmentation}\label{ap:size}
Table \ref{ap:size} shows how the MAPE performance varies when the multilingual synthetic data is further augmented through different numbers of Hindi-Marathi and Marathi-Hindi synthetic APE triplets.

\section{Additional Examples}\label{ap:add_eg}
Figure \ref{fig:add_examples_en_hi} and \ref{fig:add_examples_en_mr} show English-Hindi and English-Marathi Baseline APE and MAPE outputs pointing out some of the limitations of the current MAPE system.




\end{document}